\documentclass[twocolumn, switch]{article} 

\usepackage{preprint}

\usepackage{amsmath, amsthm, amssymb, amsfonts}
\usepackage{bm}
\usepackage{mathtools}

\usepackage[round]{natbib}
\usepackage[utf8]{inputenc}	
\usepackage[T1]{fontenc}	
\usepackage{xcolor}		
\usepackage{hyperref}
\hypersetup{colorlinks = true,
            linkcolor = blue,
            urlcolor  = blue,
            citecolor = blue,
            anchorcolor = black
}
\usepackage{booktabs} 		
\usepackage{nicefrac}		
\usepackage{microtype}		
\usepackage{lineno}		
\usepackage{float}			

\usepackage{caption}
\usepackage{array}
\usepackage[ruled,vlined]{algorithm2e}

\usepackage{newfloat}
\DeclareFloatingEnvironment[name={Supplementary Figure}]{suppfigure}

\usepackage{titlesec}
\titlespacing\section{0pt}{12pt plus 3pt minus 3pt}{1pt plus 1pt minus 1pt}
\titlespacing\subsection{0pt}{10pt plus 3pt minus 3pt}{1pt plus 1pt minus 1pt}
\titlespacing\subsubsection{0pt}{8pt plus 3pt minus 3pt}{1pt plus 1pt minus 1pt}

\usepackage{graphics}
\usepackage{multirow}
\usepackage{hhline}
\usepackage{subfigure}

\title{Predicate correlation learning for scene graph generation}

\usepackage{authblk}

\author{Leitian Tao}
\author{Li Mi}
\author{Nannan Li}
\author{Xianhang Cheng}
\author{Yaosi Hu}
\author{Zhenzhong Chen\thanks{\tt{zzchen@ieee.org}}}

\affil{School of Remote Sensing and Information Engineering, Wuhan University}

\begin{document}

\twocolumn[ 
  \begin{@twocolumnfalse} 
  
\maketitle

\begin{abstract}
	For a typical Scene Graph Generation (SGG) method, there is often a large gap in the performance of the predicates' head classes and tail classes. This phenomenon is mainly caused by the semantic overlap between different predicates as well as the long-tailed data distribution. In this paper, a Predicate Correlation Learning (PCL) method for SGG is proposed to address the above two problems by taking the correlation between predicates into consideration. To describe the semantic overlap between strong-correlated predicate classes, a Predicate Correlation Matrix (PCM) is defined to quantify the relationship between predicate pairs, which is dynamically updated to remove the matrix's long-tailed bias. In addition, PCM is integrated into a Predicate Correlation Loss function ($L_{PC}$) to reduce discouraging gradients of unannotated classes. The proposed method is evaluated on Visual Genome benchmark, where the performance of the tail classes is significantly improved when built on the existing methods.
\end{abstract}
\vspace{0.35cm}

  \end{@twocolumnfalse} 
] 



{\let\thefootnote\relax\footnote{Corresponding author: Zhenzhong Chen, E-mail:zzchen@ieee.org}}

\section{Introduction}
Scene Graph Generation (SGG) \citep{xu2017scene} is a task of interpreting an image as several object-relation triplets. Recently it has been drawing more and more attention in the computer vision community, owing to its capability of providing better middle-level feature representations in high-level tasks such as Visual Question Answering \citep{ghosh2019generating, yang2018scene}, Image Captioning \citep{yang2019auto, yang2018scene, gu2019unpaired}, Image Retrieval \citep{johnson2015image, qi2017online, ramnath2019scene}. It brings the great challenge that the objects as well as their relations should be recognized and understood simultaneously in a connected scene graph. 

Although great progress has been made in improving the feature representation of object interactions in recent years, few attention has been put on the fact that the training data of different class are highly imbalanced. On the one hand, driven by long-tailed data, most existing models are trained to ``prefer'' the \textit{head} classes\citep{tang2019learning}, which are the top-$k$ classes that have more number of data samples. Therefore, a \textit{tail} class is easily neglected and misclassified by the model. On the other hand, due to the existence of semantic overlap between similar head classes and tail classes, sometimes both classes can be regarded as correct. Such non-mutually-exclusive relation between the predicate classes twists the decision boundary and thus makes it difficult for the model to distinguish one from the other. For instance, in the Visual Genome dataset \citep{krishna2017visual}, head class ``near'' has 50 times more number of training samples than its semantically similar tail class ``on back of''. Using imbalanced dataset for training, the performance of ``near'' is almost 57 times better than ``on back of'' under the same evaluation metric. Due to the imbalanced data distribution and the semantic overlap between head and tail classes, the performance of most SGG models on tail classes is of tend not so satisfactory. 
\begin{figure}[t]
	\centering
	\subfigure[Decision boundary between predicates with strong correlation]{
	  \includegraphics[width=\linewidth]{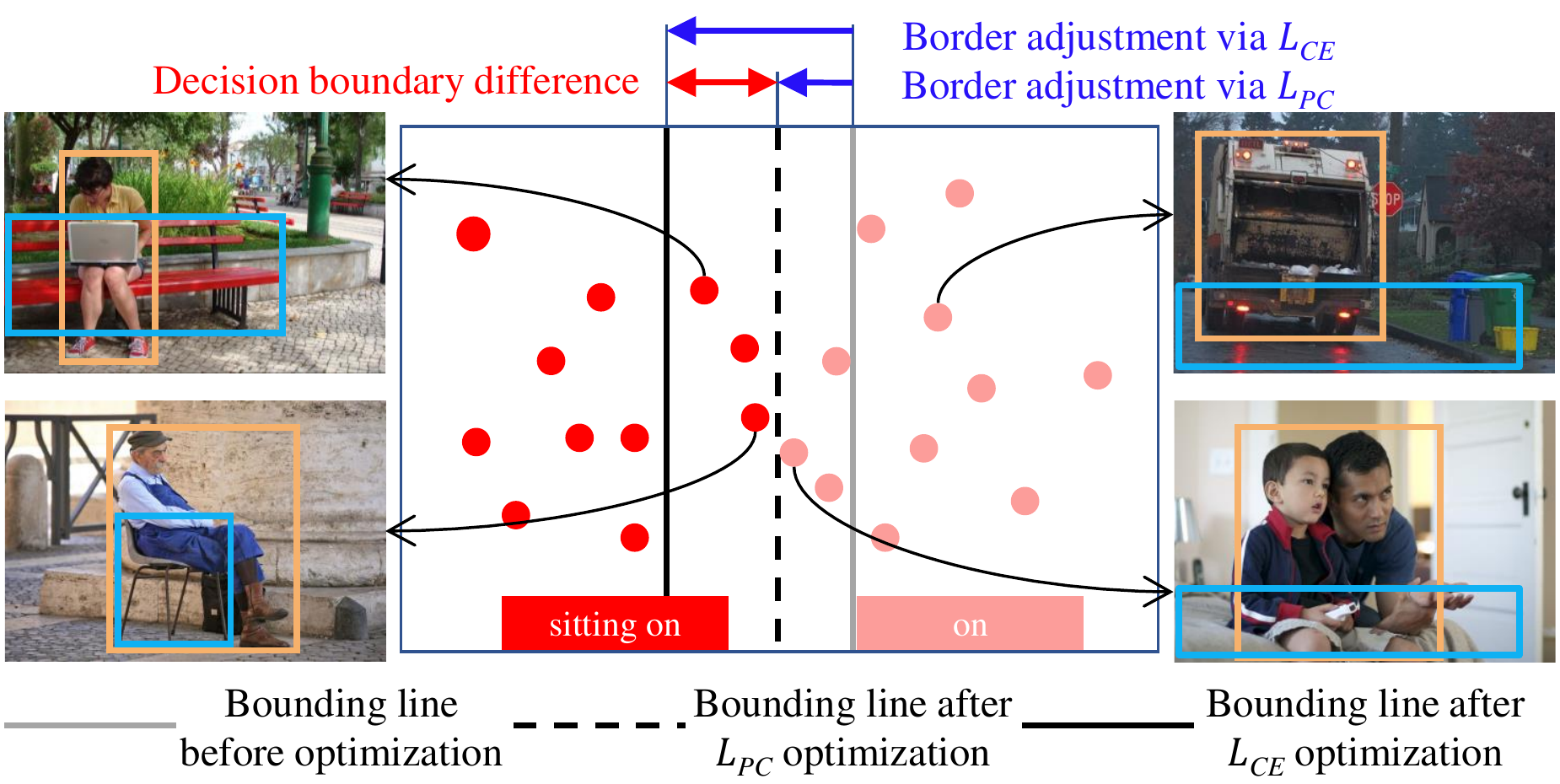}}
	\subfigure[Decision boundary between predicates with weak correlation]{
	  \includegraphics[width=\linewidth]{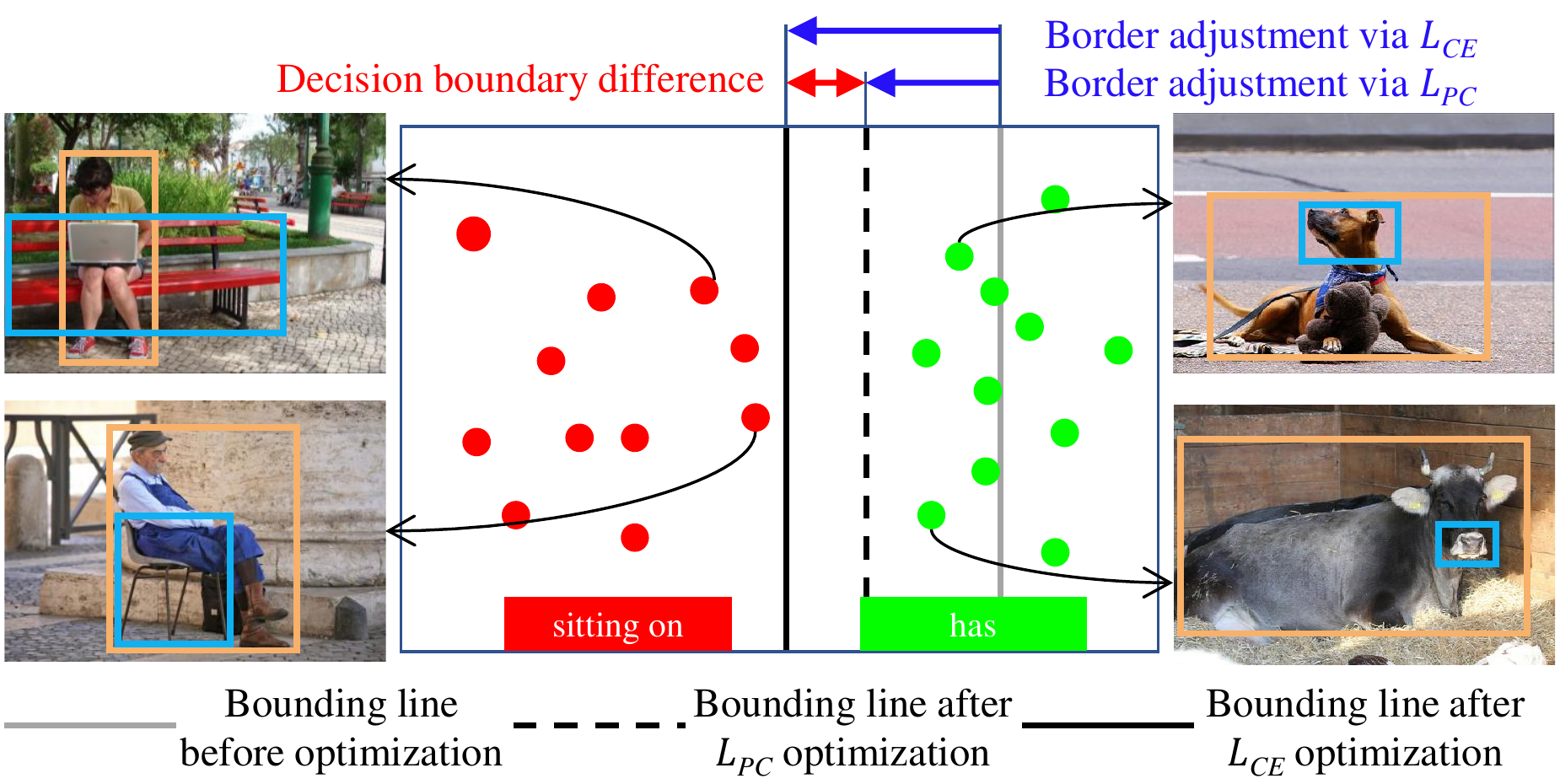}}
	\caption{The comparison between the proposed Predicate Correlation Loss function (\textbf{$L_{PC}$}), and Cross Entropy Loss function (\textbf{$L_{CE}$}) on the decision boundaries among strong correlated predicate classes (a) and weak correlated predicate classes (b). Solid and dashed lines represent the decision boundary of the pre-trained model and the decision boundary of the well-trained model, respectively.}
	\label{fig1}
\end{figure}
Aiming to promote the classification accuracy of tail classes, we put the focus on the semantic overlap between head predicates and tail predicates, and describe the overlap as the predicate correlation. To describe the semantic overlap between strong-correlated predicate classes, a Predicate Correlation Matrix (PCM) is defined to quantify the relationship between predicate pairs. To alleviate the long-tailed bias of the introduced PCM, we design an updating mechanism to refresh the correlation matrix constantly according to the current model's performance. Our predicate correlation matrix and SGG model are updated alternately to produce an accurate and less biased scene graph. Thereafter, we integrate a Predicate Correlation Loss $L_{PC}$ into the traditional cross-entropy loss function $L_{CE}$ to adjust the gradient, hoping to weaken the model's suppression for unannotated tail classes. Specifically, we quantify the predicate correlation using a well pre-trained SGG model itself. For an unannotated predicate class that is strongly correlated with the ground-truth class, the suppression for its gradient can be largely weakened such that it has a higher chance of being correctly predicted by the model. As shown in Figure \ref{fig1}, some samples from the tail class ``sitting on'' were partially predicted as the head class ``on'' when optimized with the traditional $L_{CE}$. As shown in Figure~\ref{fig1}(a), after modifying the gradient with our proposed predicate correlation loss $L_{PC}$, the decision boundary is properly set. As shown in Figure~\ref{fig1}(b), the relationship between ``has'' and ``sitting on'' is relatively weak, so the adjustment degree of $L_{PC}$ is relatively small, it doesn't affect the well-performed results adopting $L_{CE}$. 

The rest of this paper is organized as follows. Section
II gives a review of the related work. In Section III, we illustrate the proposed predicate correlation learning for scene graph generation. Section IV presents the experiments on Visual Genome. Finally, Section V concludes the paper.

%
%
%

\section{Related Work}
\paragraph{Scene Graph Generation. } SGG is to generate a visually-grounded scene graph that most accurately correlates with an image. Most of the early methods \citep{lu2016visual,yu2017visual,dai2017detecting,yang2018graph,li2018factorizable,mi2020hierarchical,xu2017scene,li2017scene, ren2020scene, hung2020contextual, guo2021relation} focused on better modeling the interactions among objects. Xu \emph{et al.} \citep{xu2017scene} firstly proposed a SGG task, and introduced a message passing mechanism to improve the relationship representation. Dai \emph{et al.} \citep{dai2017detecting} designed a framework for exploiting the statistical dependencies between objects and their relationships to tackle the problem of the ambiguous relationship. Zellers \emph{et al.} \citep{zellers2018scenegraphs} presented analysis on regularly appearing substructures in scene graphs and designed a new architecture to capture such repeated structures, as well as firstly pointed out the imbalance of predicates in the SGG dataset. Chen \emph{et al.} \citep{chen2019knowledge}, and Tang \emph{et al.} \citep{tang2019learning} noticed the bias caused by the imbalance of the SGG dataset and proposed the less biased metric: mean recall@K. They both improve the method of relation modeling to alleviate bias. Chen \emph{et al.} \citep{chen2019knowledge} proposed a routing mechanism to propagate messages through the graph to explore the statistical correlations between objects. Tang \emph{et al.} \citep{tang2019learning}  proposed to compose dynamic tree structures that place the objects in an image to capture visual contexts. Due to the effectiveness of these relation modeling methods, the scene graph's relationship representation learning has made remarkable achievements. However, due to the long-tailed distribution of the dataset, tail classes' performance is often not so satisfactory, part of the work begins to pay attention to the impact of predicates' features on relationship classification. Tang \emph{et al.} \citep{tang2020unbiased} introduced causal analysis to solve bias in the training process. Yan \emph{et al.} \citep{yan2020pcpl} proposed to judge the semantic independence of each predicate to re-weight loss. The more independent the semantic is, the greater the impact of the predicate is on the loss. Yu \emph{et al.} \citep{yu2020cogtree} designed a tree structure for the predicate class according to word semantics. While they didn't notice the semantic overlap between predicates. A method to quantitatively construct the relationship between predicates according to the pre-trained model is introduced in our work. A loss function based on the correlation of predicate to adjust the optimization direction is proposed. 
\indent
\paragraph{Imbalanced Data Distribution. }  Long-tailed data distribution is not rare in the real world scenario. A small number of classes constitute the vast majority of the data samples. If the classifier is trained using the long-tailed data directly, the performance of head class is superior, whereas the performance of tail class is not satisfactory. Mainstream methods for addressing long-tail imbalance distribution include re-sampling, and re-weighting, which re-balance the contribution of each sample from each class during the training dynamic.

Re-sampling refers to the method of obtaining samples whose frequencies are different from those of the original distribution. Generally, there are two ways of re-sampling, including under-sampling of head classes \citep{han2005borderline,chawla2002smote,singh2018clustering,byrd2019effect} and over-sampling of tail classes\citep{jeatrakul2010classification,tahir2012inverse}. However, each predicate class's images are highly diverse, so it is difficult to find the reasonable sampling strategies.

Re-weighting is to assign different weights according to the number of each class for different samples. Huang \emph{et al.} \citep{huang2016learning}, and Wang \emph{et al.} \citep{wang2017learning} tried to weight samples based on the inverse of class frequency. Lin \emph{et al.} \citep{lin2017focal} proposed Focal Loss to improve the model's training effect by alleviating the weight of easy example during training. Khan \emph{et al.} \citep{khan2019striking} proposed a new network structure based on Bayesian uncertainty to extend the classification boundaries of tail classes. Cui \emph{et al.} \citep{cui2019class} solved the problem of data re-ensembling by introducing the concept of adequate sample number. Wu \emph{et al.} \citep{wu2020distribution} focused on multi-label classification under the long-tail distribution and proposed strategies to solve the problem of re-balance weighting negative-tolerant regularization. Tan \emph{et al.} \citep{tan2020equalization} redesigned softmax to protect the learning of rare classes. 

Beyond that, some specific learning strategies are also used to address long-tail distribution, such as transfer learning \citep{liu2020deep}, metric learning \citep{huang2016learning} and meta-learning \citep{shu2019meta}. Two-stage training approach is also an effective way. Kang \emph{et al.} \citep{kang2019decoupling} found that the distribution of image features and class distribution is not coupled, so feature learning and classifier parameter should be updated separately. Based on the conclusion of \citep{kang2019decoupling}, Li \emph{et al.} \citep{li2020overcoming} enhanced samples of hard classes by group softmax. 

\section{Predicate Correlation Learning For Scene Graph Generation}
We first provide an overview of current approaches for scene graph generation based on the standard cross entropy loss, followed by a description of our proposed predicate correlation learning  method for SGG.

\begin{figure*}
  \centering
  \includegraphics[width=\linewidth]{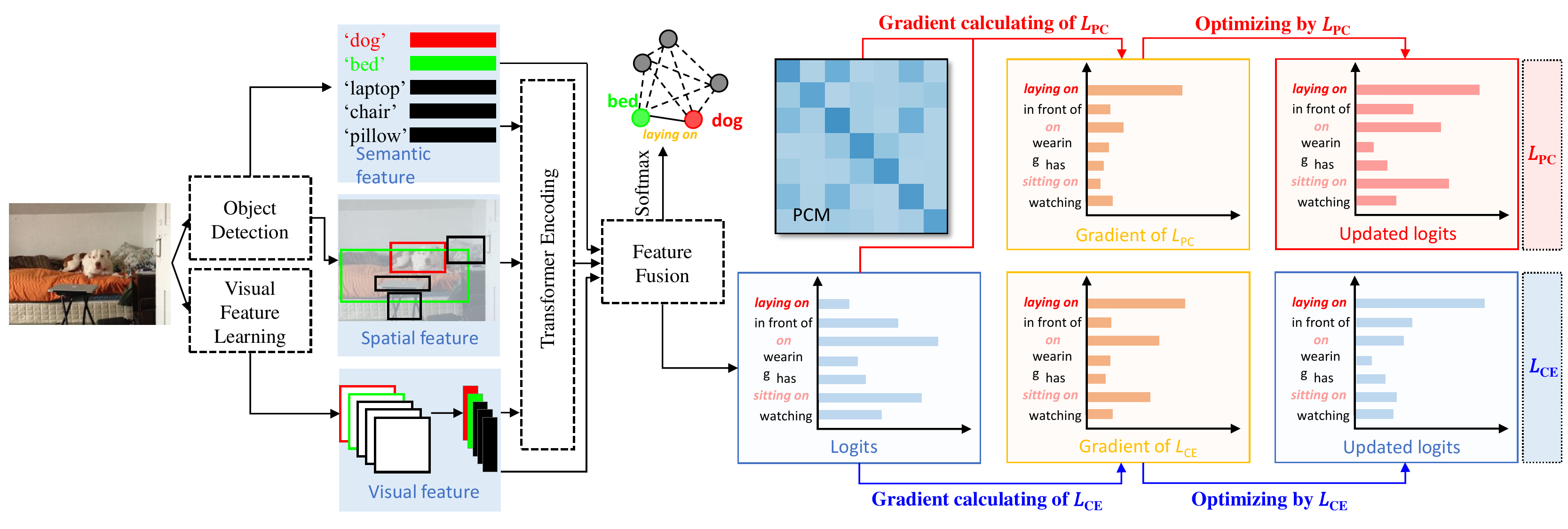}
  \caption{Overview of Predicate Correlation Learning method. The proposed method equipped with a standard feature representation module. The traditional methods with $L_{CE}$ directly adopt the softmax layer to classify the relationship, and then calculates the gradient according to the logits and then adjusts the parameters. Our $PCL$ with $L_{PC}$ considers the correlation between predicates. Using $PCM$ and logits to calculate the gradient together, the proposed method weakens the suppression of strongly correlated classes. By $L_{PC}$, unannotated predicate classes' probabilities are higher after optimization compared to $L_{CE}$.} 
  \label{fig2}
\end{figure*}
\subsection{Scene Graph Generation Framework}
As shown in the Figure~\ref{fig3}, SGG methods typically adopt a two-stage framework, consisting of two modules: visual feature representation learning and relationship classification.  

In the first stage, the SGG model gets the kinds of features needed for relationship classification and forms the graph's nodes. Given an image $I$, a set of bounding boxes $B=\left\{b_{1}, b_{2}, \ldots, b_{n}\right\},$ representing the position of the detected objects on the image, are obtained using a Faster R-CNN \citep{ren2016faster} object detector. Features corresponding to these regions are extracted by RoIAlign, which is is a kind of operation that transforms the area of arbitrary size into a standard size by bilinear interpolation. According to the spatial position on the image represented by $B$, a set of object features $R=\left\{r_{1}, r_{2}, \ldots, r_{n}\right\}$ are obtained. The next step is to determine the object class of $b_{i}$. A list of object class labels $L=\left\{l_{1}, l_{2}, \ldots, l_{n}\right\},$ consistent with the $B$ are predicted. 
\begin{equation}
  \text { Input: }\left\{\left(R, B, L \right)\right\} \Longrightarrow \text { Output }:\left\{O\right\}
\end{equation}
These features form the nodes of the scene graph, and the nodes are defined as $O=\left\{o_{1}, o_{2}, \ldots, o_{n}\right\}$. 

In the second stage, visual feature representations are used as inputs to predict edge (relationship between objects) between nodes in the scene graphs.
As shown in the left of Figure \ref{fig2}, according to the summary in \citep{tang2020unbiased}, when judging the edge between node $i$ and node $j$, it can be divided into the following three input feature to classify the relationship between objects:
\begin{itemize}
  \item Object Feature Input ( $o_{e}$): 
  \begin{equation}
    \text { Input : }\left\{o_{i}, o_{j}\right\} \Longrightarrow \text { Output : }\left\{o_{e}\right\}
  \end{equation}
  Visual object feature plays an important role in distinguishing the type of objects and predict the relation between them. There may be quite different visual relationships between the same pair of objects. ``\textit{man - standing on - surfboard}'' and ``\textit{man - carrying - surfboard}'' are different in poses.
  \item Visual Context Input ($v_{e}$): 
  \begin{equation}
  v_{e}= Convs \left(\operatorname{RoIAlign}\left(\mathcal{M}, b_{i} \cup b_{j}\right)\right) 
  \end{equation}
  where $b_{i} \cup b_{j}$ indicates the union box of two RoIs, $\mathcal{M}$ indicates the feature map of image $I$. The spatial feature is a meaningful feature for relationship prediction. For example, according to the relative position of bounding box, we can judge whether the predicate between two objects is more likely to be ``in front of'' instead of ``on back of''.
  \item Object Class Input ($l_{e}$): 
  \begin{equation}
    \text { Input : }\left\{l_{i}, l_{j}\right\} \Longrightarrow \text { Output : }\left\{l_{e}\right\}
  \end{equation}
\end{itemize}

To a large extent, there may be a relatively fixed relationship between two specific objects. This inference depends on bias of the language instead of the visual feature.
Finally, the prediction of the predicate is determined by the above three features after fusion. The fusion function is as follows:
\begin{equation}
  p_{e} = W_{o} o_{e}+W_{v} v_{e}+W_{l} l_{e}
\end{equation}

The last part of the model is the softmax layer, which inputs features $p_{e}$ and adopts softmax to calculate the probability of each relationship. Most of previous work adopt cross-entropy as the loss function of relationship classification. Each relationship is only assigned one ground-truth label. Under this formulation, the classification of predicates is regarded as a single-label classification problem, while the relationship between the same pair of objects can be described by a variety of predicates. In this work, we propose a Predicate Correlation Loss to take the correlation of predicates into consideration. 
\subsection{Predicates Correlation Matrix}
By analyzing the misclassification of tail classes in a biased model, we design a method to construct $PCM$, and then a mechanism is designed to alleviate bias in $PCM$.
\subsubsection{Analysis of misclassification of the biased model}
SGG methods always performs well on head predicate classes, but poor on tail predicate classes. We analyze why tail classes have poor performance. First of all, we find that they are often wrongly predicted as head classes such as ``on/near/has'', while these predicates are not rich in semantics and less helpful for downstream tasks. This is due to the extremely unbalanced distribution of samples. In addition, it is found that misclassification mainly occurs in the following cases: 1) There are semantically overlapping predicates, such as ``has'' and ``with''. 2) Some predicates share similarity in the statuses they represent , such as ``standing on'' and ``walking on''. 3) Some fine-grained classes can be categorized into coarse-grained classes that own a large number of samples, such as ``walking on'' is often predicted as ``on''. It can be concluded that due to the previous efforts in feature representation learning, the existing SGG models can distinguish the semantically unrelated classes well, while the misclassification mainly occurs between the classes that have strong correlation as explained above. Since the expression of the relationship between objects is not unique, these misclassifications may not be considered as errors for incomplete annotation (when prediction is ``sitting on'' while ground-truth label is ``on''). To some extent, the high misclassification rate of tail classes may be partly attributed to the single class annotation for the relationship between objects. 

\subsubsection{Definition}
Firstly, we need to obtain the results of a well-trained model on the validation. Record the probabilities of each predicate output under the triplet of the same ground-truth. \textit{$P_{ij}^{k}$} means the logit as follows:
\begin{equation}
  P_{ij}^{k}= \mathcal{F}(x = j | Predicate=i)
\end{equation}
where the $i$ and $j$ both denote the index of predicate class, the upper subscript $k$ denotes the index of the sample, $\mathcal{F}$ denotes function of model prediction.
After normalization, the probability of all $P_{i}$ under $j$ is averaged, which is $PCM_{i j}$, Since the logit values of the output are relatively small, Softmax Scaling is too smooth to make enough difference between different predicate correlations. We adopted calculating the ratio for each P and then take the means.
\begin{equation}
PCM_{i j}=\frac{1}{N} \sum_{k=1}^{N} \frac{P_{i j}^{k}}{\sum_{j=1}^{n}P_{ij}^{k}},
\end{equation}
where $N$ denotes the total number of predicate\ $i$'s samples, $n$ denotes the number of predicate class.
\subsubsection{Updating Mechanism}
According to the previous analysis, the initial model is biased whether it is learning directly from the long-tailed dataset or re-weighted data. The initial parameters of $PCM$ are from a biased SGG model, so $PCM$ is also biased inevitably. Therefore, the correlation matrix should be constantly updated to alleviate the initial model's long-tailed bias. A mechanism is designed to update the parameters utilizing the accurate predicate relationship instead of bias from the dataset, thus guiding the model to optimize in the right direction. 

Since each epoch will be tested on the validation set, a new $PCM$ can be calculated after each epoch. Because our method alleviates the bias by reducing unreasonable suppression, the new model always has a smaller bias than the previous model. By calculating the new $PCM$ and the original $PCM$, we can get a $PCM$ with a smaller bias 
\begin{equation}
  {PCM}_n={\mu PCM}_{n-1}+{(1-\mu) PCM}_k
\end{equation}
where $PCM_{k}$ denotes the new matrix from the new model, $\mu$ denotes proportional speed of $PCM$ updating, and $PCM_{n}$ represents the value of $PCM$ after $n-1$ updates.
\subsection{Predicate Correlation Loss Function}
We first review the traditional cross entropy loss function and then introduce the proposed predicate correlation loss function.
\subsubsection{Cross Entropy Loss Function}
As one of the most widely used loss functions in classification, the form of cross-entropy loss ($L_{CE}$) can be presented as follows:
\begin{equation}
  L_{CE}=-\sum_{i=1}^{C} y_{i} \log \left(p_{i}\right) \quad, \quad \text { with } p_{i}=\frac{e^{z_{i}}}{\sum_{j=1}^{C} e^{z_{j}}}
  \label{ce}
\end{equation}
where $z_{i}$ denotes the logit of the i-th class predicted by the model, $p_{i}$ denotes the normalized probability calculated by $\sigma(z)$, $y$ adopts one-hot representation. $\sum_{j=1}^{C} y_{j}=1$, and $y_{t} = 1$, which means class $t$ is the ground-truth predicate. $L_{CE}$ calculates the cross entropy between the estimated distribution \textbf{\emph{p}} and the true distribution \textbf{\emph{y}}. When calculating the gradient of $L_{CE}$, the formulation is as follows:
\begin{equation}
  \frac{\partial L_{CE}}{\partial z_{i}}=\left\{\begin{array}{ll}
    p_{i}-1, & \text { if } i = t \\
    p_{i}, & \text { if } i \neq t
    \end{array}\right.
\end{equation}
As shown in Eq.(\ref{ce}), for a foreground sample of class $t$, it can be regarded as a negative sample for any other class $f (f\neq t)$. The class $f$ will receive a discouraging gradient $p_{f}$ during the model's parameters updating, which will lead the network to predict a low probability for class $f$. However, if there is a strong correlation between $t$ and $f$, the discouraging gradients are unreasonable because $f$ may also be an acceptable expression for this pair of objects. The accumulated unreasonable and discouraging gradients have a non-negligible impact on the learning of that predicate class.

\subsubsection{Predicate Correlation Loss Function}
During training, the relationship between predicates needs to be considered to adjust the optimization direction. In this work, we propose the Predicate Correlation Loss Function ($L_{PC}$) that allows the relationship classification module of SGG models taking the correlation of predicates into consideration. The traditional equalization loss \citep{tan2020equalization} can be adopted to update the gradient of back-propagation to avoid some unreasonable suppression. To make it more effective to the semantic overlap in SGG, we introduce PCM to the traditional equalization loss, and name it as  Predicate Correlation Loss ($L_{PC}$). By reducing the suppression of predicate that has a strong correlation of ground-truth predicate according to PCM, our $L_{PC}$ optimizes the model's parameters that can be presented as follows:
\begin{equation}
  L_{P C}=-\sum_{i=1}^{C} y_{i} \log \left(\tilde{p_{i}}\right), \quad \text { with } \tilde{p_{i}}=\frac{e^{z_{i}}}{\sum_{j=1}^{C} w_{i j}e^{z_{j}}}
  \label{eq2}
\end{equation}
where $w_{i j}$ denotes the weight of predicate class $j$ when the sample predicate is $i$, which can be presented as:
\begin{equation}
  w_{i j}=1-\left(1-y_{i}\right) pcm_{i j},\quad pcm_{i j}\in (0,1)
\end{equation}
Here $pcm_{i j}$ indicates the correlation between $i$ and $j$. The stronger the correlation between $i$ and $j$ is, the larger $pcm_{i j}$ is. According to the weight of $w_{ij}$, the absolute value of loss can be adjusted. Due to the existence of the $1-y_{i}$ coefficient, for the ground-truth class $t$, $w_{it}$ is 1. For the class $f (f\neq t)$, the stronger the correlation between $t$ and $f$ is, the smaller $w_{tf}$, and thus $\tilde{p_{t}}$ is smaller when compared to the $p_{t}$ calculated by $L_{CE}$. Therefore, in this case, the extent of parameters' adjustment is smaller than $L_{CE}$. When the predicate $f$ and ground-truth predicate $t$ is almost unrelated, $w_{tf}$ is close to 1, the impact on the parameter update margin
is negligible. According to the above analysis, the weight $w_{ij}$ can effectively adjust the parameter update margin on the basis of the correlation between predicates. When calculating the gradient of $L_{PC}$ back propagation, the formulation is as follows:
\begin{equation}
  \frac{\partial L_{P C}}{\partial \tilde{z_{i}}}=\left\{\begin{array}{c}
    \tilde{p_{i}}-1, \quad \quad \text { if } i=t \\
    \left(1-pcm_{i j}\right) \tilde{p_{i}}, \text { if } i \neq t
    \end{array}\right.
\end{equation}
Unlike the gradient of cross entropy shown in Eq.(\ref{eq2}), the gradient of our $L_{PC}$ back propagation takes the correlation between predicates classes into consideration. For the discouraging gradient of predicate $f (f\neq t)$, the gradient is adjusted according to the correlation between $t$ and $f$. The margin of the gradient change is positively correlated with the correlation between predicate $f$ and ground-truth $t$. Therefore, the unreasonable suppression of the strongly correlated class is weakened according to the correlation. If there is almost no relationship between $i$ and $j$, the back propagation gradient is almost unchanged. The proposed loss can optimize the model in a more reasonable direction by reducing unreasonable suppression for strong-related predicates while keeping the gradient of independent predicates. By modifying the gradient with $L_{PC}$, the decision boundary is better set compared to $L_{CE}$. The comparison between our $L_{PC}$ and the cross entropy loss in parameter updating and gradient adjustment is shown in the right part of Figure \ref{fig2}.

On the other hand, due to the long-tail distribution of predicate classes, it is clear that the model will have severe bias without re-balancing. During the network parameter updating, rare classes are at a disadvantage due to the overwhelming, discouraging gradient, so their performance is unsatisfactory. It is unwise to judge the relationship between predicates based on such a biased model. We introduced a classical re-balancing strategy to alleviate the bias, so a less biased $PCM$ is obtained. We set the samples of each class with different importance and weight them according to the number of each class sample. Under the strategy of re-weighting, the formulation of the $L_{PC}^{*}$ is as follows:
\begin{equation}
  L_{PC}^{*}=-W_{k} \sum_{i=1}^{C} y_{i} \log \frac{e^{z_{i}}}{\sum_{j=1}^{C} w_{i j} e^{z_{j}}}
\end{equation}
Class Balanced Loss \citep{cui2019class} is a well performed re-weighting strategy. The sample weights are estimated based on effective numbers. We adopt this weighting factor:
\begin{equation}
  W_{k}=\frac{1-\beta}{1-\beta^{n_{k}}}
\end{equation}
where $\beta$ denotes the hyper-parameter that represents the sample domain, $n_{k}$ denotes the sample number of predicate \ $k$.

The proposed Predicate Correlation Learning method adopts $L_{PC}^{*}$ as the loss function. Since the ground-truth label denotes as input when PCL is adopted, which is unknown during inference, we use the softmax for multi-classification as a surrogate.
\begin{table*}[t]
  \centering
  \caption{Mean recall comparison with state-of-the-art methods on VG150 dataset. The constrained mR@20/50/100 in $\%$ on PredCls, SGCls and SGDet tasks are presented.}
  \scalebox{0.79}{
    \begin{tabular}{cccccccccccc}
    \toprule
    \multirow{2}[4]{*}{Backbone} & \multirow{2}[4]{*}{Model} & \multirow{2}[4]{*}{Methods} & \multicolumn{3}{c}{Predicate Classification} & \multicolumn{3}{c}{Scene Graph Classification} & \multicolumn{3}{c}{Scene Graph Detection} \\
\cmidrule{4-12}          &       &       & mR @ 20 & mR @ 50 & mR @ 100 & mR @ 20 & mR @ 50 & mR @ 100 & mR @ 20 & mR @ 50 & mR @ 100 \\
    \midrule
    \multirow{7}[2]{*}{VGG} & IMP+  & -     & -     & 9.8   & 10.5  & -     & 5.8   & 6.0   & -     & 3.8   & 4.8  \\
          & Motif & -     & 10.8  & 14.0  & 15.3  & 6.3   & 7.7   & 8.2   & 4.2   & 5.7   & 6.6  \\
          & KERN  & -     & -     & 17.7  & 19.2  & -     & 9.4   & 10.0  & -     & 7.1  & 9.8  \\
          & VCTree & -     & 14.0  & 17.9  & 19.4  & 8.2   & 10.1  & 10.8  & 5.2   & 6.9   & 8.0  \\
          & GPS-Net & -     & 17.4  & 21.3  & 22.8  & 10.0  & 11.8  & 12.6  & 6.9   & 8.7   & 9.8  \\
          & PCPL  & -     & -     & 35.2  & \textbf{37.8 } & -     & 18.6  & 19.6  & -     & 9.5   & 11.7  \\
          & Transformer  & PCL     & \textbf{30.9 } & \textbf{35.4 } & 37.4  & \textbf{16.5 } & \textbf{19.6 } & \textbf{20.9 } &  \textbf{8.1}     & \textbf{9.9}      & \textbf{12.4} \\
    \midrule
    \multirow{10}[6]{1.5cm}{ResNeXt-\quad101-FPN } & \multirow{4}[2]{*}{Motif} & Baseline & 11.5  & 14.6  & 15.8  & 6.5   & 8.0   & 8.5   & 4.1   & 5.5   & 6.8  \\
          &       & TDE   & 18.5  & 25.5  & 29.1  & 9.8   & 13.1  & 14.9  & 5.8   & 8.2   & 9.8  \\
          &       & CogTree & 20.9  & 26.4  & 29.0  & 12.1  & 14.9  & 16.1  & 7.9   & 10.4  & 11.8  \\
          &       & PCL  & \textbf{28.1 } & \textbf{33.6 } & \textbf{35.8 } & \textbf{15.5 } & \textbf{18.2 } & \textbf{19.1 } & \textbf{10.9 } & \textbf{14.2 } & \textbf{16.6 } \\
\cmidrule{2-12}          & \multirow{4}[2]{*}{VCTree} & Baseline & 11.7  & 14.9  & 16.1  & 6.2   & 7.5   & 7.9   & 4.2   & 5.7   & 6.9  \\
          &       & TDE   & 18.4  & 25.4  & 28.7  & 8.9   & 12.2  & 14.0  & 6.9   & 9.2   & 11.1  \\
          &       & CogTree & 22.0  & 27.6  & 29.7  & 15.4  & 18.8  & 19.9  & 7.8   & 10.4  & 12.1  \\
          &       & PCL  & \textbf{26.4 } & \textbf{32.9 } & \textbf{35.7 } & \textbf{21.7 } & \textbf{25.2 } & \textbf{26.3 } & \textbf{10.9 }      &\textbf{14.8 }       &\textbf{17.4 }  \\
\cmidrule{2-12}          & \multirow{2}[2]{*}{Transformer} & Baseline & 13.1  & 16.5  & 18.0  & 7.3   & 9.5   & 10.1  & 5.8   & 7.8   & 9.1  \\
          &       & PCL  & \textbf{33.2} & \textbf{36.3} & \textbf{39.2} & \textbf{17.9} & \textbf{20.7} & \textbf{21.8} & \textbf{11.1} & \textbf{15.2} & \textbf{18.3} \\
    \bottomrule
    \end{tabular}%
    }
  \label{table1}%
\end{table*}%
\begin{table*}[t]
  \centering
  \caption{Recall comparison with debiasing methods on VG150 dataset. The constrained R@20/50/100 in $\%$ on PredCls, SGCls and SGDet tasks are presented.}
  
  \scalebox{0.9}{
    \begin{tabular}{cccccccccc}
    \toprule
    \multirow{2}[4]{*}{Methods} & \multicolumn{3}{c}{Predicate Classification} & \multicolumn{3}{c}{Scene Graph Classification} & \multicolumn{3}{c}{Scene Graph Detection} \\
\cmidrule{2-10}          & R @ 20 & R @ 50 & R @ 100 & R @ 20 & R @ 50 & R @ 100 & R @ 20 & R @ 50 & R @ 100 \\
    \midrule
    PCPL  & -     & 50.8  & 52.6  & -     & 27.6  & 28.4  & -     & 14.6  & 18.6  \\
    Transformer+PCL  & \textbf{48.0 } & \textbf{54.3 } & \textbf{56.1 } & \textbf{28.7 } & \textbf{32.2 } & \textbf{33.1 } & \textbf{13.5 } & \textbf{18.3 } & \textbf{21.6 } \\
    \midrule
    Motif+TDE & 33.6  & 46.2  & 51.4  & 21.7  & 27.7  & 29.9  & 12.4  & 16.9  & 20.3  \\
    Motif+CogTree & 31.1  & 35.6  & 36.8  & 19.4  & 21.6  & 22.2  & 15.7  & 20.0  & 22.1  \\
    Motif+PCL & \textbf{47.0 } & \textbf{55.0 } & \textbf{57.3 } & \textbf{30.4 } & \textbf{34.2 } & \textbf{35.2 } & \textbf{22.1 } & \textbf{29.0 } & \textbf{33.4 } \\
    \midrule
    VCTree+TDE & 36.2  & 47.2  & 51.6  & 19.9  & 25.4  & 27.9  & 14.0  & 19.4  & 23.2  \\
    VCTree+CogTree & 39.0  & 44.0  & 45.4  & 27.8  & 30.9  & 31.7  & 14.0  & 18.2  & 20.4  \\
    VCTree+PCL & \textbf{44.9 } & \textbf{53.4 } & \textbf{56.2 } & \textbf{34.4 } & \textbf{38.4 } & \textbf{39.5 } & \textbf{21.5 } & \textbf{27.6 } & \textbf{31.9 } \\
    \midrule
    Transformer+PCL & \textbf{50.5 } & \textbf{57.3 } & \textbf{59.2 } & \textbf{33.2 } & \textbf{36.0 } & \textbf{37.0 } & \textbf{23.0 } & \textbf{29.9 } & \textbf{34.2 } \\
    \bottomrule
    \end{tabular}%
  }
  \label{table2}%
\end{table*}%

\section{Experiments}
\subsection{Experiment Settings}
\subsubsection{Implementation Details}
To keep consistency with previous works, we adopt the Faster-RCNN \citep{ren2016faster} as object detector, pre-trained on ImageNet \citep{russakovsky2015imagenet} and refined on VG150 \citep{krishna2017visual}, with VGG16 \citep{simonyan2014very} and ResNeXt-101-FPN \citep{lin2017feature, xie2017aggregated} being the backbone to generate region proposals. When Motif and VCTree are adopted as the baseline, the initial learning rate is 0.01. When transformer is adopted as the baseline, the initial learning rate is 0.001. All the experiments are implemented with PyTorch and conducted with NVIDIA 1080 GPUs. $\beta$ is adopted as 0.9999.
\subsubsection{Dataset}
In order to keep up with previous work \citep{xu2017scene,zellers2018scenegraphs,tang2019learning,chen2019knowledge,tang2020unbiased,yan2020pcpl}, we experiment on the same Visual Gemome \citep{krishna2017visual} dataset that has been cleaned and sorted. We use the most frequent 150 object classes and  predicates for evaluation. As a result, each image has a scene graph of around 11.5 objects and 6.2 relationships. Following the previous work, we use the same dataset division, in which 5000 images are selected as validation, with 70$\%$ of the images for training, and the remaining 30$\%$ for testing.
\subsubsection{Tasks}
We test the methods in three different sub-tasks\citep{xu2017scene}:
\begin{itemize}
  \item Predicate Classification (PredCls): Given the ground-truth annotations of the object classes and bounding boxes, predict each object pair's relation type.
  \item Scene Graph Classification (SGCls): Given the ground-truth annotations of object bounding boxes, predict the object classes and each object pair's relation type.
  \item Scene Graph Generation (SGDet): Predict the bounding boxes, the object classes, and the relation type of each object pair.
\end{itemize}

\subsubsection{Metrics}
Following \citep{chen2019knowledge,tang2019learning}, considering that the distribution of relationships is highly imbalanced in VG, we utilize mean recall@K (mR@K) as the main metrics. mR@K retrieves each predicate separately and then averages R@K for all predicates

\subsection{Comparison with State-of-the-art Methods}
In this section, firstly, we compare with other biased scene graph generation methods, including IMP+ \citep{xu2017scene}, Motif \citep{zellers2018scenegraphs}, KERN \citep{chen2019knowledge}, VCTree\citep{tang2019learning} and GPS-Net \citep{lin2020gps}. These methods mainly focus on better feature representation learning networks. Due to the imbalanced data distribution and the semantic overlap between head and tail classes, tail classes' performance is often not so satisfactory, As shown in the top of Table \ref{table2}, the mean recall of these methods is relatively low. If the bias of the dataset is not alleviated, the mean recall can not be so satisfactory due to the poor performance of tail classes.

\begin{table*}[t]
  \centering
  \caption{Ablation study for loss function and matrix adopted by our $L_{PC}$. The constrained mR@20/50/100 in $\%$ on PredCls, SGCls and SGDet tasks are presented.}
  \scalebox{0.855}{
    \begin{tabular}{c|cc|cccccccccccc}
    \hline
    \multicolumn{1}{c}{\multirow{2}[4]{*}{Type}} & \multicolumn{5}{c}{Methods}           & \multicolumn{3}{c}{Predicate Classification} & \multicolumn{3}{c}{Scene Graph Classification} & \multicolumn{3}{c}{Scene Graph Detection} \\
\cmidrule{7-15}    \multicolumn{1}{c}{} & $L_{PC}$ & $L_{PC}^{*}$ & PCM   & PCM(r) & PrM   & mR@20 & mR@50 & mR@100 & mR@20 & mR@50 & mR@100 & mR@20 & mR@50 & mR@100 \\
    \hline
    1     & \checkmark     &       &       &       &       & 13.1  & 16.5  & 18.0  & 7.3   & 9.5   & 10.1  & 5.8   & 7.8   & 9.1  \\
    2     & \checkmark     &       &       & \checkmark      &      & 13.3  & 16.4  & 18.1  & 7.2   & 9.4   & 10.3  & 5.7   & 8.1   & 9.4  \\
    3     & \checkmark     &       &       &     & \checkmark       & 14.0  & 17.9  & 19.4  & 8.2   & 10.1  & 10.9  & 6.3   & 8.4   & 9.9  \\
    4     & \checkmark     &       & \checkmark     &       &       & 14.8  & 18.8  & 20.4  & 8.8   & 10.5  & 11.4  & 6.8   & 8.9   & 10.2  \\
    \hline
    5     &       & \checkmark     &       &       &       & 29.1  & 34.5  & 36.3  & 15.5  & 18.4  & 19.5  & 9.7   & 14.0  & 16.5  \\
    6     &       & \checkmark     &       &     & \checkmark       & 29.4  & 34.7  & 36.9  & 15.8  & 18.9  & 19.8  & 10.1  & 14.3  & 17.0  \\
    7     &       &\checkmark       &     & \checkmark       &     & 29.6  & 36.3  & 39.2  & 17.9  & 20.7  & 21.8  & 11.1  & 15.2  & 18.3  \\
    \hline
    \end{tabular}%
  }
  \label{table3}%
\end{table*}%

\begin{table}[h]
  \centering
  \caption{Ablation study for $\mu$. The constrained mR@20/50/100 in $\%$ on PredCls task are presented.}
    \begin{tabular}{cccc}
    \hline
    \multirow{2}[2]{*}{$\mu$} & \multicolumn{3}{c}{Predicate Classifications} \\
          & mR@20 & mR@50 & mR@100 \\
    \hline
    0     & 29.9  & 35.6  & 37.9 \\
    0.5   & \textbf{30.2}  & 36.1  & 38.5 \\
    0.9   & 29.6  & \textbf{36.3}  & \textbf{39.2} \\
    \hline
    \end{tabular}%
  \label{table4}%
  \vspace{0.5em}
\end{table}%

We perform a more in-depth comparison between our method and several debiasing strategies in SGG to demonstrate our method's effectiveness further. Our method is evaluated on three baseline models: Motif, VCTree, and Transformer. No debiasing strategy is used in the baseline. We compare the performance with the other debiasing approach: PCPL \citep{yan2020pcpl}, TDE \citep{tang2020unbiased} and CogTree \citep{yu2020cogtree}. All the hyper-parameters settings are consistent with the previous work. We compare our methods with PCPL, a method that re-weights data according to the degree of independence of predicates. As shown in Table \ref{table1} and Table \ref{table2}, in the PredCls setting, our recall is 6.0$\%$ higher than that of PCPL, although our mR@100 and PCPL are similar. For SGCls and SGDet setting, the mR@100 of our method exceeds PCPL 6.0$\%$ and 6.0$\%$, and the R@100 of our method is 16.5$\%$ and 16.1$\%$ higher than PCPL. This proves that compared with PCPL, our method not only improves the performance of tail predicates but also maintains the performance of head predicates. Motif and VCTree have been experimented as the baseline of TDE and CogTree. We also compare them with our method under the same experimental baseline and setting. Take Motif+TDE as an example: under three sub-tasks, our mR@100 is 18.3$\%$, 28. 2$\%$ and 69. 4$\%$ higher than TDE, respectively. Meanwhile, our R@100 is 11.5$\%$, 17.7$\%$, and 54.7$\%$ higher than TDE, respectively. As for CogTree + VCTree, the comparison with our $PCL$ is as follows : under three sub-tasks, our mR@100 is 20. 2$\%$, 32. 2$\%$ and 43. 8$\%$ higher than CogTree, respectively. Meanwhile, our R@100 is 23.8$\%$, 24.6$\%$, and 56.4$\%$ higher than CogTree, respectively. It can be seen that our method has been dramatically improved on the three baselines, which proves the effectiveness of our method.

\subsection{Ablation Study}
All ablation experiments are equipped with the Faster R-CNN with a ResNeXt-101-FPN backbone.
\subsubsection{Analysis of loss function}
As shown in Table \ref{table3}, we mainly conduct ablation experiments under two settings: cross entropy ($L_{CE}$) used as loss function training and class balanced loss \citep{cui2019class} ($L_{CB}$) used for re-balancing training to alleviate bias. Different matrixes are used to fuse into the Predicate Correlation Learning. $PCM$ is a matrix describing the predicate correlation obtained without the re-balancing strategy, and $PCM(r)$ represents the less biased matrix obtained under the re-weighting strategy. In order to prove the effectiveness of the $PCM$, we set the probability matrix ($PrM$), which defines the probability of a class appearing in a dataset as the degree of correlation with other classes, as a comparative experiment. According to the comparison between type 1 and type 2, it is found that the introduction of $PrM$ into our $L_{PC}$ doesn't have much effect on improving mean recall. According to the comparison between type 1 and type 3, we find that using $PCM$ in $L_{PC}$ is indeed helpful to the performance of the tail class. Compared with adopting $L_{CE}$ directly and using our loss, we find that using $PCM$ in $L_{PC}$ can improve the performance of the tail class. In the three sub-tasks, the distribution of mR@100 increase by 7.7$\%$, 7.9$\%$, and 8.8$\%$. 

Furthermore, we carry on a re-weighting experiment to alleviate the impact of bias from the dataset. As shown in the comparison between 5 and 7, under the strategy of re-weighting, in the three sub-tasks, the distribution of mR@100 increase by 8.0$\%$, 11.8$\%$ and 11.0$\%$. As shown in the comparison between type 1 and type 4, it can be seen that the long-tail distribution does seriously affect the learning of tail classes. As an effective strategy of re-balancing, class balance loss really alleviates the bias of datasets and improves the performance of tail classes. Therefore, it is very likely that the tail class will be predicted into a small class without re-balancing, resulting in a severe bias in constructing a predicates correlation matrix. Therefore, re-balancing to alleviates bias is really necessary for the definition of $PCM$. Then we migrate the $PCM$ obtained under the re-weighting strategy to the long tail dataset to prove the effect of re-weighting on modeling $PCM$. As shown in the comparison between type 3 and type 4, under the strategy of re-weighting, in the three sub-tasks, the distribution of mR@100 increase by 5.2$\%$, 4.6$\%$ and 3.0$\%$. 
\begin{figure*}
  \centering
  \includegraphics[width=\linewidth]{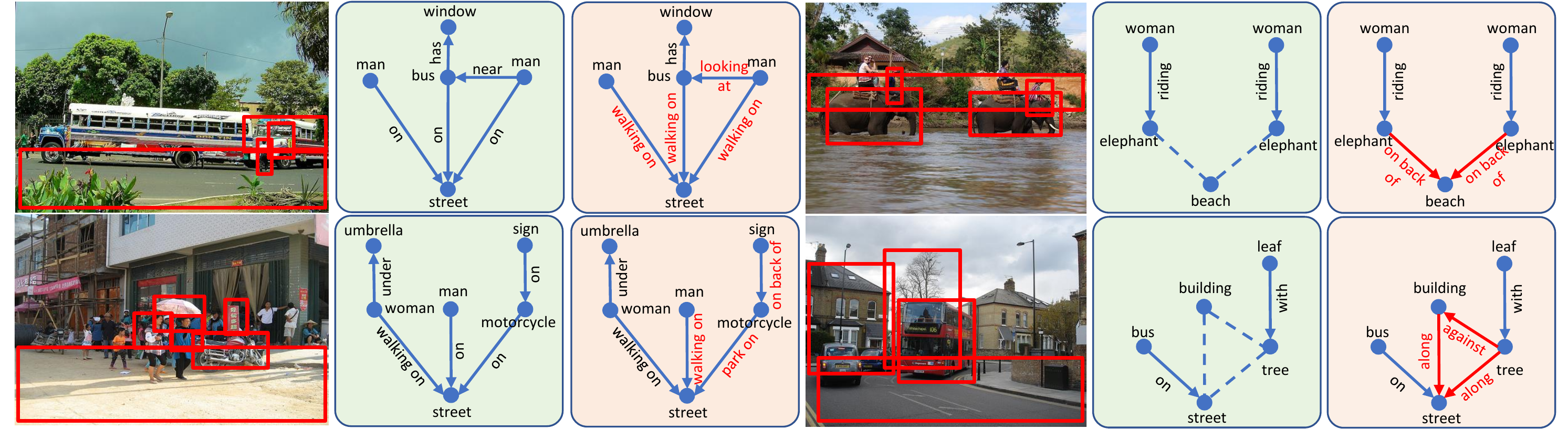}
  \caption{Visualizations of scene graphs generated by a transformer model. For each image, the left scene graph is generated by the model trained using cross-entropy loss (in green) and the right scene graph is generated by the model trained using proposed \textbf{$PCL$} (in red).
  } 
  \label{fig3}
\end{figure*}
\begin{table*}[t]
  \centering
  \caption{Qualitative analysis of different learning strategies. The mR@100 and of the variance of norms under different strategies are shown, and we list the performance of predicates with different sample sizes (top m-n).}
  
    \begin{tabular}{cccccccc}
    \hline
    \multirow{2}[4]{*}{Methods} & \multirow{2}[4]{*}{Variance of norm} & \multirow{2}[4]{*}{mR@100} & \multicolumn{5}{c}{mR@100 for top m-n predicates} \\
\cmidrule{4-8}          &       &       & 1-10  & 11-20 & 21-30 & 31-40 & 41-50 \\
    \hline
    $L_{CE}$ & 0.0154 & 18.0  & 54.9  & 19.2  & 12.0  & 3.3   & 0.5  \\
    $L_{CE+PC}$ & 0.0151 & 19.4  & \textbf{55.0 } & \textbf{21.5 } & 12.9  & 6.5   & 1.2  \\
    $L_{CB}$ & 0.0111 & 36.3  & 47.1  & 35.8  & 37.8  & 29.5  & 31.3  \\
    $L_{CB+PC}$ & \textbf{0.0101} & \textbf{39.2 } & 48.1  & 38.3  & \textbf{41.3 } & \textbf{32.3 } & \textbf{36.0 } \\
    \hline
    \end{tabular}%
    \vspace{1.5em}
  \label{table5}%
\end{table*}%
\subsubsection{Analysis of $\mu$}
In order to judge the effect of different update parameters on the performance of the model, we take different values of $\mu$. We set three values in ablation of $\mu$: 0, 0.5 and 0.9. $\mu=0$ represents the $PCM$ obtained by using the bias model directly without updating. $\mu=0.5$ represents the $PCM$ updated at a fast rate, and $\mu=0.9$ represents the $PCM$ updated at a slow rate. As shown in Table \ref{table4}, we can observe that if the parameters are not updated, the original model's performance is not as good as updating to alleviate the bias. The update speed of $PCM$ smoother can achieve better results.
\subsection{Qualitative Analysis}
We visualize several PredCls samples that are generated by the baseline and our $PCL$ in Figure \ref{fig3}. Some relationships in the baseline method are predicted to be coarse-grained predicates due to the semantic overlap between different predicates as well as the long-tailed data distribution. Through our $PCL$, the relationship between objects is successfully judged to be more meaningful and fine-grained predicate. It shows that our $PCL$ greatly improves the baseline method.

According to \citep{kang2019decoupling} and \citep{li2020overcoming}, the classification layer's class weight norms are imbalanced. The weight norms of predicate positively correlate with the number of training samples. Tail predicate classes get few chances to be activated and are suppressed constantly, so the weight norms of tail classes are smaller than that of head classes. Such imbalanced classifiers (parameter norm) would make the classification scores for tail classes much smaller than those of head classes. Therefore, the model is biased and tends to predict the relationship as the head class. As shown in Table \ref{table5}, we calculate and analyze the performance and norm of head class and tail class under different strategies. If the re-weighting strategy is adopted, the situation is just the opposite. The class with small weight is always suppressed to a great extent, which leads to the smaller norm. The difference of different predicate norms in classification layers can reflect the bias degree of the model. As shown in Table \ref{table5}, the variance of the model's norm trained by our method is smaller through analysis and calculation. This means the model with our $PCL$ has less bias. It proves that our method successfully weakens the unreasonable suppression in parameter updating and makes the optimization direction more reasonable. Our method does alleviate the bias caused by unreasonable annotations of datasets instead of over fitting the tail class to get higher mean Recall. 

We conduct a more in-depth analysis of the performance of the learning strategies under the four methods. We calculate the average norm and mR@100, and the average R@100 of five groups of predicates of different numbers. As shown in Table \ref{table5}, we have made a more detailed analysis of the performance of each group of class with different number of samples under different learning strategies. 
Different loss represents different strategies. $L_{CE}$ means learning directly on the unbalanced dataset. $L_{CE+PC}$ means to consider class correlation. $L_{CB}$ means to solve the long-tail problem through heavy weighting, and $L_{CB+PC}$ means to handle the problem of semantic overlap as well as the problem of long-tail distribution. It can be seen that considering the relationship between predicates is helpful to improve the performance of the head class and the tail class.
\section{Conclusions}
Due to the semantic overlap between different predicates as well as the long-tailed data distribution, the predicate's tail class is often misclassified to its semantically similar head class. To alleviate the problem, we take the introduced correlation between predicates into consideration and present a $PCL$ method. The proposed method adjusts the gradient according to the correlation matrix, which guides the optimization into properer direction. Simultaneously, an updating mechanism is introduced to remove the matrix's long-tailed bias. Moreover, this method is built on various SGG models and proven to improve the performance drastically, which shows the effectiveness of our approach.


\newpage
\normalsize
\bibliography{myBib}


\end{document}